\documentclass[conference,compsoc]{IEEEtran}
\ifCLASSOPTIONcompsoc
\usepackage[nocompress]{cite}
\else
\usepackage{cite}
\fi
\ifCLASSINFOpdf
\else
\fi

\usepackage{float}
\usepackage{graphicx}
\usepackage{amssymb,amsmath,amsthm,amscd}
\usepackage{subfigure}

\usepackage{multirow}
\usepackage{array}
\makeatletter
\newcommand{\thickhline}{%
	\noalign {\ifnum 0=`}\fi \hrule height 1.1pt
	\futurelet \reserved@a \@xhline
}
\newcolumntype{"}{@{\hskip\tabcolsep\vrule width 1pt\hskip\tabcolsep}}
\makeatother

\begin{document}
	%
	\title{High-Frequency aware Perceptual Image Enhancement}

	\author{\IEEEauthorblockN{Hyungmin Roh}
		\IEEEauthorblockA{The Interdisciplinary Program of\\
			Computational Science and Technology\\
			Seoul National University\\
			Seoul, Republic of Korea\\
			raingold1347@snu.ac.kr}
		\and
		\IEEEauthorblockN{Myungjoo Kang}
		\IEEEauthorblockA{Department of Mathematics\\
			Seoul National University\\
			Seoul, Republic of Korea\\
			mkang@snu.ac.kr}}

	
	%


	\maketitle
	
	\begin{abstract}
		In this paper, we introduce a novel deep neural network suitable for multi-scale analysis and propose efficient model-agnostic methods that help the network extract information from high-frequency domains to reconstruct clearer images. Our model can be applied to multi-scale image enhancement problems including denoising, deblurring and single image super-resolution. Experiments on SIDD, Flickr2K, DIV2K, and REDS datasets show that our method achieves state-of-the-art performance on each task. Furthermore, we show that our model can overcome the over-smoothing problem commonly observed in existing PSNR-oriented methods and generate more natural high-resolution images by applying adversarial training.
	\end{abstract}
	

	%
	\IEEEpeerreviewmaketitle

	\section{Introduction}
	
	Most learning-based methods utilize the high capacity of deep neural networks with remarkable ability to understand the content and style of the image that they have shown in visual recognition tasks, including image classification and object detection. Using these high capacities and analytic powers of deep neural networks, learning-based methods have been successfully adapted to the field of image enhancement and have shown better performances compared to traditional model-based methods in laboratory environments.

	When applied to real-world problems, however, most learning-based methods have failed to produce such good results while model-based methods are more flexible and applicable to low-resolution images with various kinds of blur and noises. This is because learning-based methods learn how to enhance the quality of images only by analyzing relations between given pairs of low-resolution images and their corresponding high-resolution ones in the training phase. However, in real-world problems, only low-resolution images are given and their high-resolution pairs are unknown. This means that the models have to infer new relations that they have never learned, which often leads to huge performance degradation when they solve real-world problems.

	Another problem called the ``ill-posed problem'' also makes solving real-world problems more challenging; there are countless high-resolution image candidates in solution spaces corresponding to a given low-resolution image, while the number of high-resolution outputs human viewers perceive natural is very small or unique. The ill-posed problem makes it very difficult for deep neural networks to derive natural high-resolution outputs when solving real-world problems. Research on mathematical ways to reduce the solution spaces in unsupervised environments has been recently proposed to deal with the problem.

	Many studies on the architectural design of deep neural networks have been proposed over the years and have shown great performances. However, they have recently reached the limit; little progress has been made except for marginal improvements on performances. This is because deep neural networks are originally optimized for understanding the content of images based on the high capacity of deeply stacked layers, so they are less capable of interpreting and restoring detailed information of corrupted images. Accordingly, recent studies are more focused on conveying mathematical properties of images to the existing models rather than designing deeper networks.

	In keeping with this trend, we not only propose novel architectures of deep neural network for image enhancement problems but also introduce some state-of-the-art model-agnostic methods to make networks capable of producing sharper and more realistic images by providing abstract characteristics and high-frequency components of images with a little modification in the structure of existing models.

	\section{Related work}
	
	\begin{figure*}[ht]
		\centering
		\includegraphics[width=.9\textwidth]{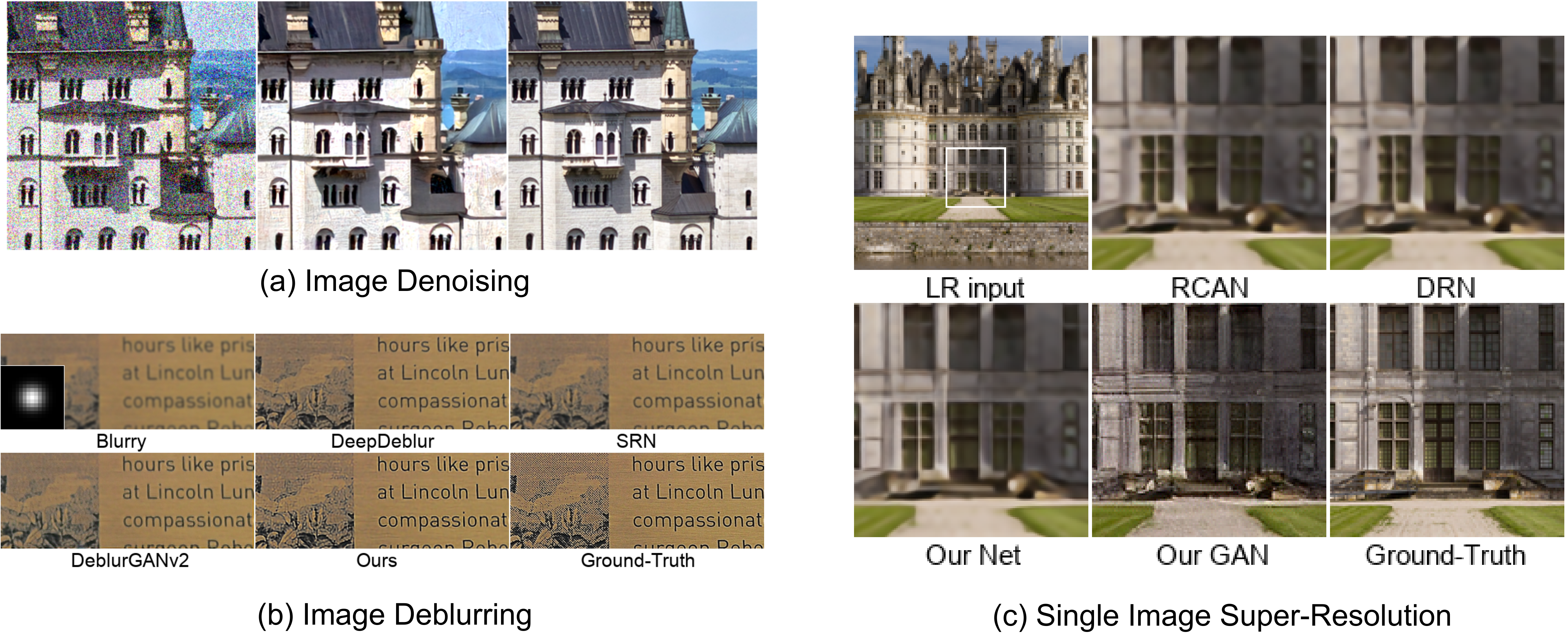}
		\caption{Examples of visual results of our Image Enhancement Method}\label{fig:result}
	\end{figure*}
	
	In recent years, many studies have been proposed to solve the SISR problem using different deep-learning techniques. In 2015, Dong et al.~\cite{srcnn} introduced deep learning methods into the SISR problem, proposing SRCNN that is a fully convolutional neural network that enables end-to-end mapping between input and output images. In 2016, Kim et al.~\cite{vdsr} proposed VDSR that utilizes contextual information spread over large patches of images using large receptive fields to convolutional layers. In 2017, Tai et al.~\cite{drrn} proposed a very deep network structure consisting of 52 convolutional layers called DRRN by designing a recursive block with a multi-path structure while Ledig et al.~\cite{srresnet} proposed SRResNet with 16 blocks of deep ResNet and also introduced GAN-based SRGAN which is optimized for perceptual loss calculated on feature maps of the VGG~\cite{vgg} network.
	
	In 2017, Lim et al.~\cite{edsr} proposed a novel model named EDSR. They removed every batch normalization from their network and stacked 16 residual blocks, which extracts high-frequency information from low-resolution images. In the same year, Tong et al.~\cite{srdensenet} proposed SRDenseNet, which consists of 8 dense blocks~\cite{densenet} and skip connections that combine feature maps from different levels. In 2018, Zhang et al.~\cite{rdn} introduced a residual dense block that allows direct connections from preceding blocks, leading to a continuous memory mechanism. Zhang et al.~\cite{rcan} also proposed a novel model called RCAN, which added channel attention to EDSR and introduced a Residual in Residual module to construct a 10 times deeper network. They used skip connections with various lengths to help their model separately extract abundant low-frequency features and scarce but important high-frequency information from low-resolution images.
	
	Until 2019, studies have mainly focused on modifying networks' architectural design by introducing or combining various kinds of neural blocks. However, as the neural networks became sufficiently deep and wide, structural modifications alone could expect nothing but only small marginal improvements. To overcome such issues, researchers have recently focused on the intrinsic limitations of the SISR problem or attempted to combine their neural networks with traditional model-based methods.
	
	In 2020, Guo et al.~\cite{drn} introduced cycle consistency to their network to solve the intrinsic ill-posed problem; there are infinite high-resolution images that can be downsampled to the given low-resolution input images. They reconstructed the RCAB proposed by RCAN~\cite{rcan} into a UNet~\cite{unet} structure. In this process, they also produced images with $1/2$ and $1/4$ size of the target resolution from the low-resolution inputs, and then compared them with downscaled output images. Through this process, which is named dual regression, they could maintain cycle consistency and enable their networks trained with unlabeled data at the same time. Pan et al.~\cite{physics_sr} constrained their network with input information by utilizing a pixel substitution scheme from low-resolution images. They added degraded image blurred by known blur kernel to the input image and forwarded them iteratively into the deblurring network. From this process, they tried to convert a given difficult blind kernel problem to an easy non-blind problem so that their model can restore sharp images more easily.
	
	Instead of solving the ill-posed problem by giving cycle consistency to the network with constraint from input information, several attempts have been proposed to create a human interpretable network structure by applying meaningful kernel to the convolutional layers of the network. Huang et al.~\cite{defian} introduced a Multi-Scale Hessian Filtering (MSHF) consisting of kernels that extract edges from multi-scale, leading their model to approach the high-frequency information of images from different angles and scales. On the other hand, Shang et al.~\cite{rfb_esrgan} uses rectangular-shaped receptive fields such as $1\times 3$ or $3\times 1$ in parallel rather than randomly initializing $3\times 3$ convolutional kernels. In this way, their model, named RFB-ESRGAN, becomes human interpretable and could adaptively analyze both horizontal and vertical information of images.
	
	\begin{figure*}[ht]
		\centering
		\includegraphics[width=.9\textwidth]{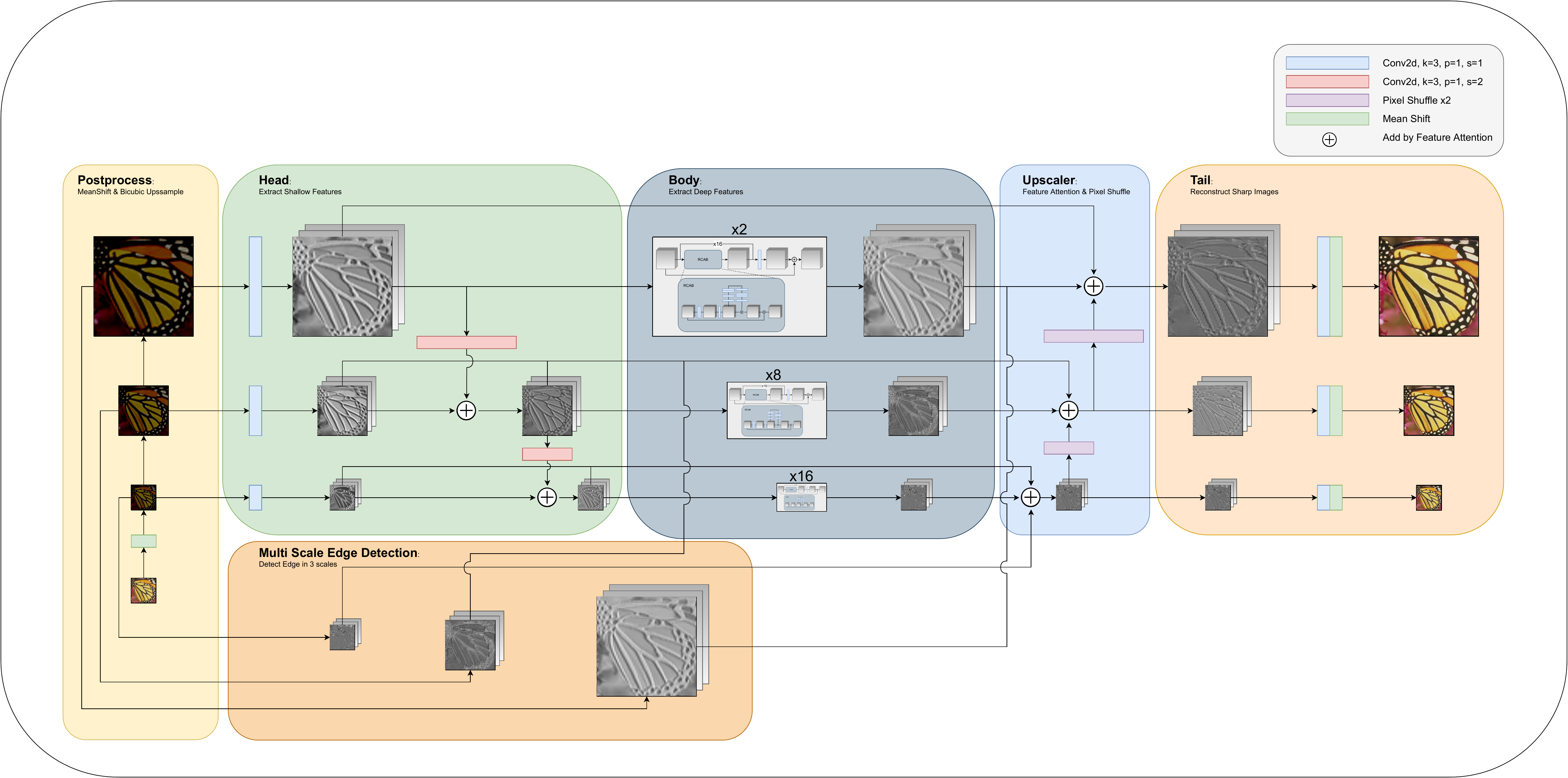}
		\caption{Our proposed network}\label{fig:network}
	\end{figure*}
	
	A study has also been proposed to apply knowledge distillation to the SISR problem to enable models to use the rich information in high-resolution images during the training phase. Lee et al.~\cite{pisr} forward the encoded feature of HR images to the teacher network, which shares the same structure as the student network, allowing the teacher network to use privileged information to obtain better outcomes. The student network then used variational information distillation~\cite{vid} technique that allows the teacher network to distill their encoded features to the student network so it can learn how to extract privileged information, allowing the model to extract more meaningful features from a given low-resolution input.
	
	As EDSR~\cite{edsr} and RCAN~\cite{rcan} separately extract shallow and deep features from the image on RGB color space, a study that tried to take a step further from color domain to frequency domain and decompose high frequency and low-frequency information has been proposed. Pang et al.~\cite{fan} split input images into high, medium, and low frequencies and passed them to the network individually, and then aggregated each convoluted feature map adaptively to generate high-resolution images. However, instead of using mathematical methods such as FFT or DWT, they simply divided the frequency domain using three convolutional layers, which is easy to fail to extract valid and meaningful frequency information.
	
	\section{Our proposed method}

	\subsection{Multi-scale Edge Filtering}
	
	\begin{figure}[ht]
		\centering
		\includegraphics[width=.4\textwidth]{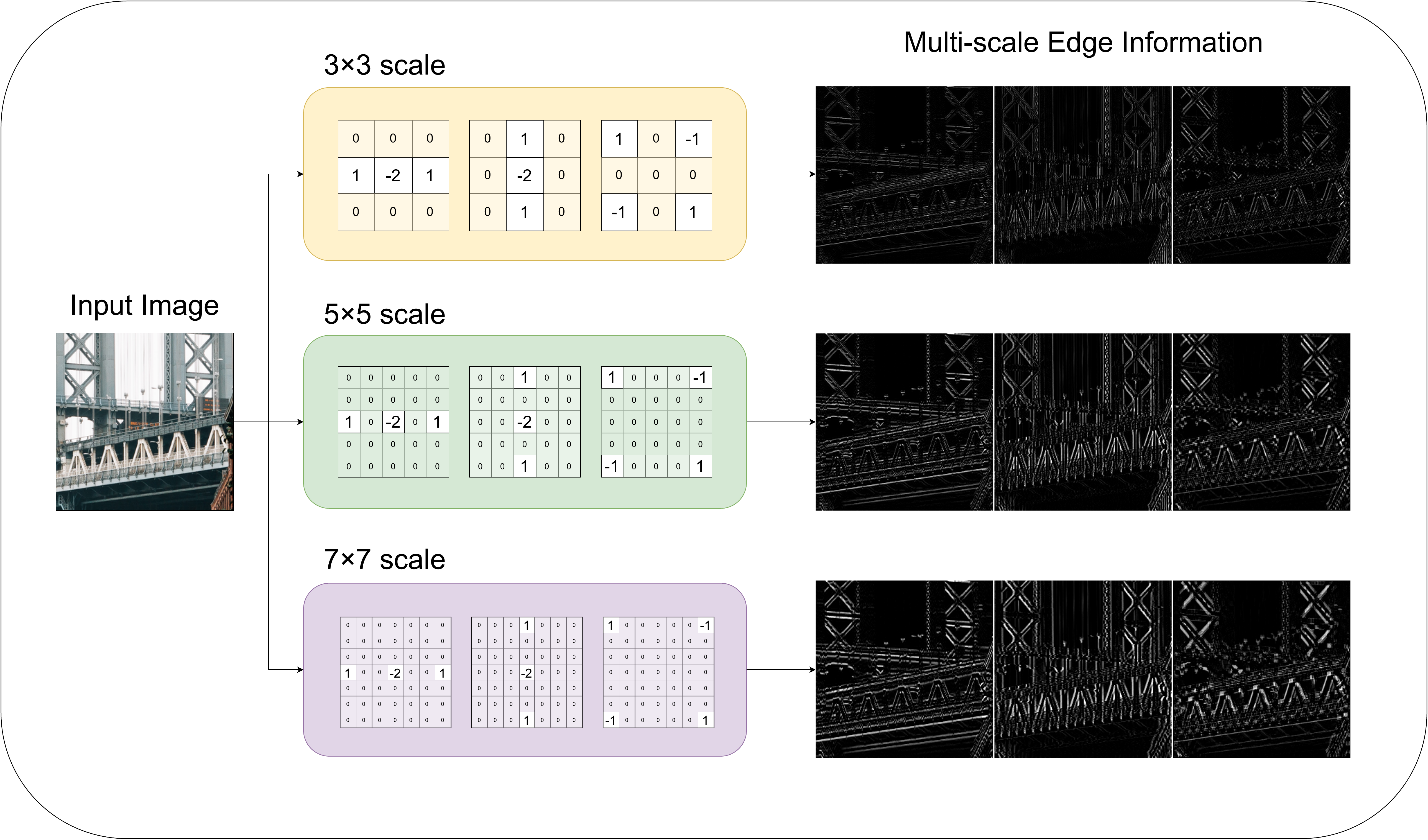}
		\caption{Multi-scale Edge Filtering}\label{fig:mshf}
	\end{figure}
	
	Successful super-resolution requires the understanding of structure of images. In particular, we need to separate high-frequency and low-frequency regions and make adaptively appropriate analyses for each region to successfully detach the noise map from the original image. This is because the distribution of pixel value appears different in each region; pixels in high-frequency regions often have large variance while smaller variances are more observed in low-frequency areas.
	
	We propose a module that extracts edges from given images to obtain information about high-frequency areas. The obtained information is transferred to the network and used to increase restoring performance by focusing more on high-frequency regions that are difficult to reconstruct. The module consists of convolutional layers initialized with pre-defined filters, making the back-propagation scheme possible and enabling end-to-end optimization when the network is training the data.

	\subsection{Feature Attention Module}\label{sec:feamodule}
	
\begin{figure}[h]
		\centering
		\subfigure[Channel Attention]{\includegraphics[width=.45\textwidth]{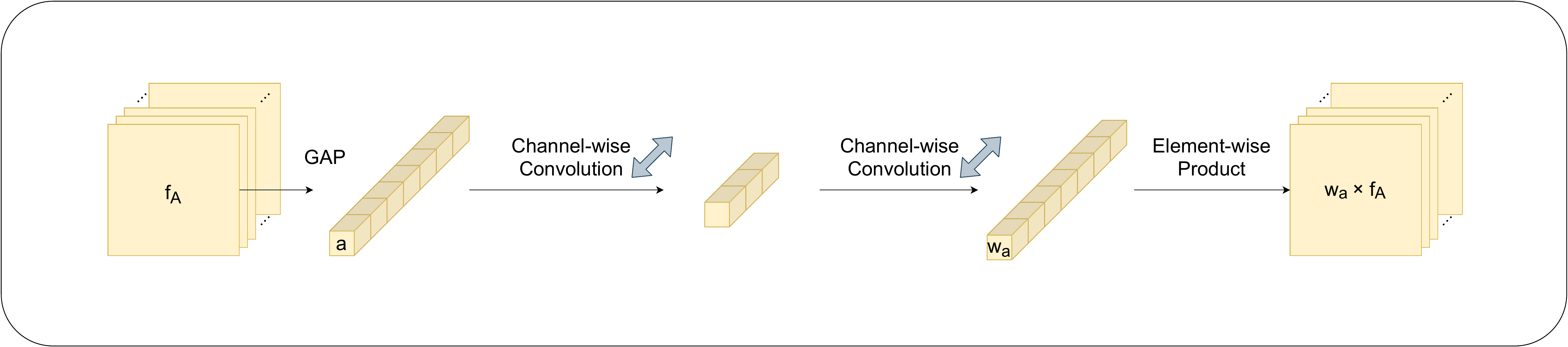}}
		\subfigure[Feature Attention]{\includegraphics[width=.45\textwidth]{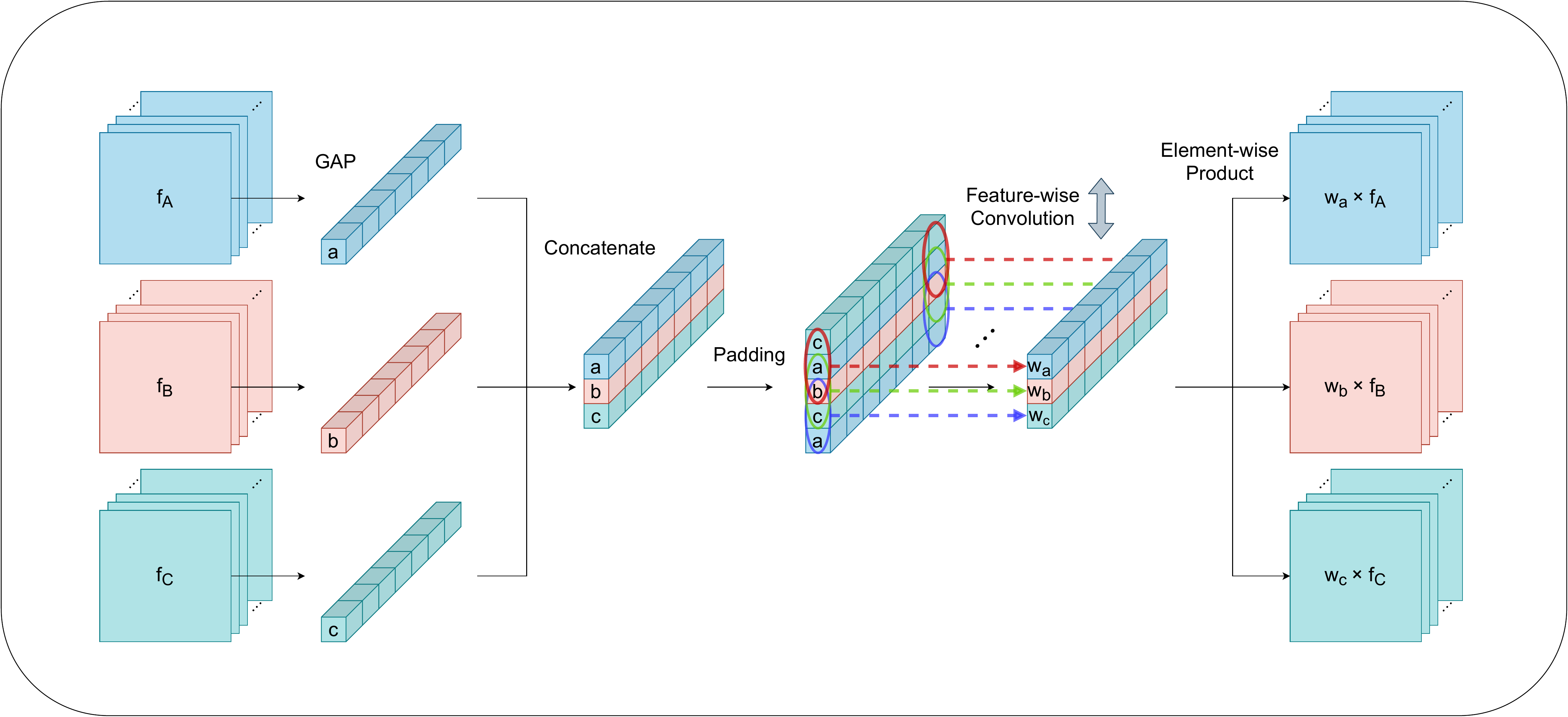}}
		\caption{Feature Attention Module}
		\label{fig:feaAtt}
	\end{figure}

	RCAN~\cite{rcan} achieved better results by adding channel attention to residual blocks from EDSR~\cite{edsr}. Figure~\ref{fig:feaAtt} (a) illustrates the concept of channel attention. The channel attention takes a vector pooled from feature maps as input and feed-forward it through a series of convolutional layers. Here, the layers give us weights for each channel by operating dot products for local channel-wise regions from the average pooled vector. This process allows the network to determine the importance between channels in the feature map and focus on channels with more information.
	
	EDSR and RCAN restore images using feature maps obtained by summing the shallow features and deep features. However, as they simply added two features, they have failed to consider the relative importance of shallow and deep features. Since shallow and deep features contain different kinds of information, such as low and high-frequency, their importance cannot be the same. Also, the characteristic of given image changes which feature contains more information. Therefore, it is necessary to introduce a module that identifies the characteristics of given images and determines each importance before adding feature maps with different information.
	
	To solve this problem, we introduce the Feature Attention Module. Before feature maps are added, the relative weights of importance are estimated by our Feature Attention Module. We calculated weight of importance in a vector form with the dimension of the channel in feature maps, considering that each channel has different importance.
	
	Figure~\ref{fig:feaAtt} shows the structure of our feature attention module. First, we concatenate vectors from each feature map by using the global average pooling layer. Then we pad the stack of vectors and feed them into convolution in the feature-wise direction rather than the channel-wise way. We padded the vectors to maintain the output dimension and make the convolution to compute every feature evenly. By multiplying each feature map in an element-wise way, we could finally obtain a weighted sum of features depending on their importance.
	
	\subsection{High-Pass Filtering Loss}
	
	\begin{figure}[h]			
		\centering			
		\subfigure[]{			
			\includegraphics[width=.15\columnwidth]{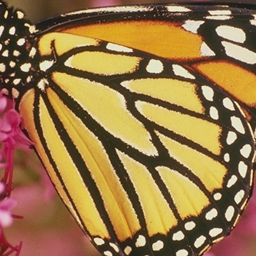}				
			\label{fig:freq_a}				
		}			
		\subfigure[]{				
			\includegraphics[width=.15\columnwidth]{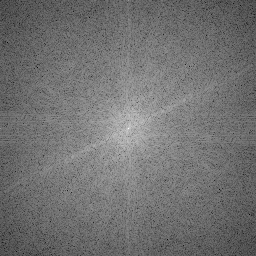}		
			\label{fig:freq_b}				
		}
		\subfigure[]{			
			\includegraphics[width=.15\columnwidth]{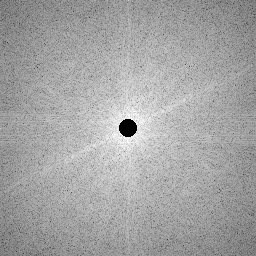}				
		}
		\subfigure[]{				
			\includegraphics[width=.15\columnwidth]{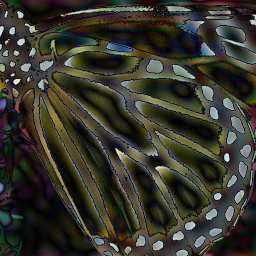}				
		}
		\subfigure[]{				
			\includegraphics[width=.15\columnwidth]{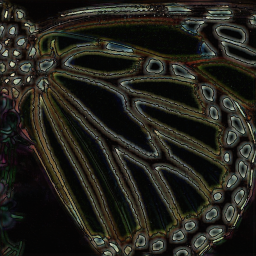}				
		}
		\caption{A visual example of high-pass filtering. (a) Original image. (b) Frequency spectrum in the polar form where the spectrum is shifted to place zero frequency at the center. (c) High-pass filtered Frequency spectrum. (d) High-pass filtered image, or inverse Fourier transform of (c). (e) High-frequency domain of original image from our model.}
		\label{fig:highpass_filter}
	\end{figure}
	
	\begin{figure}[h]
		\centering
		\includegraphics[width=.4\textwidth]{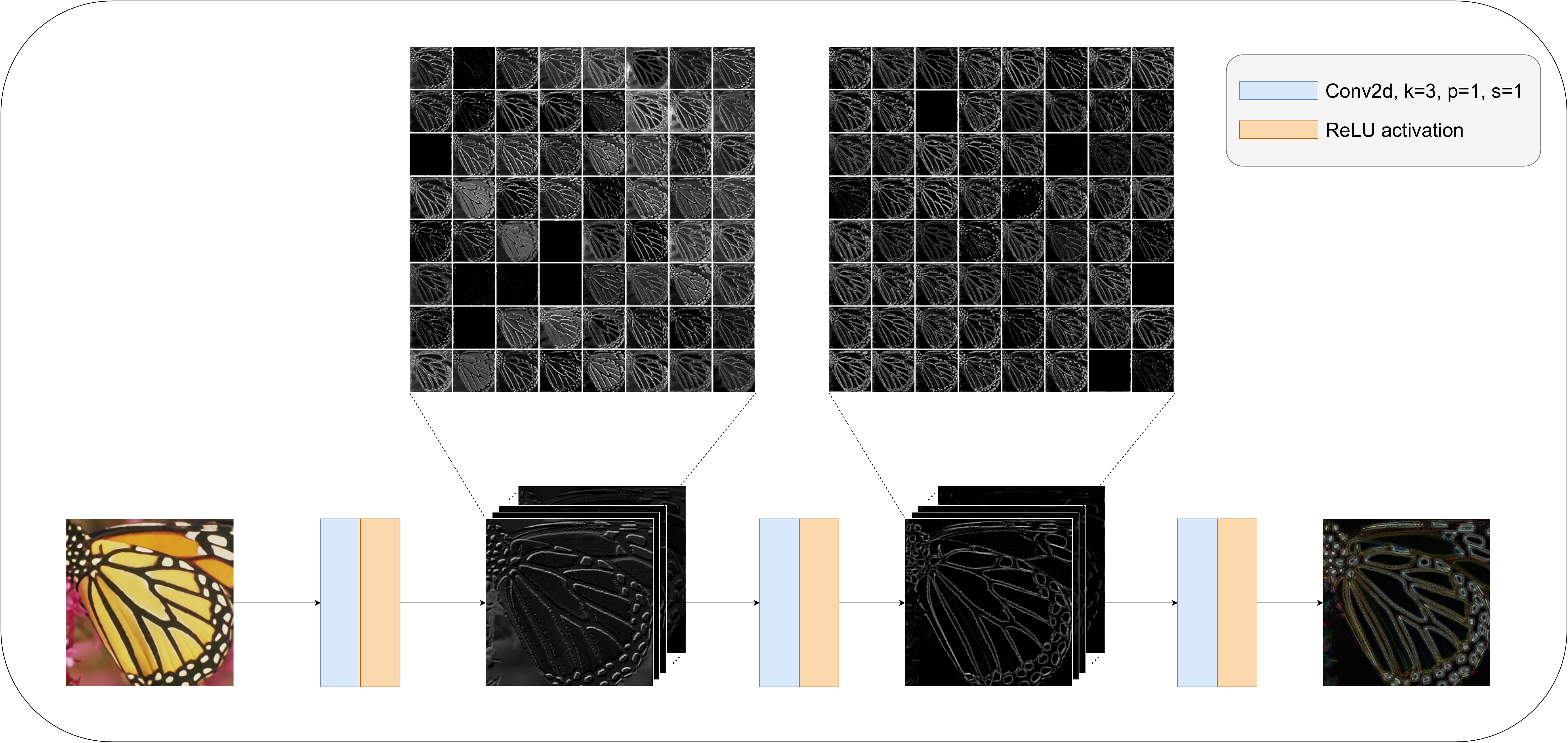}
		\caption{High-Pass Filtering with CNN Model}
		\label{fig:highpass_model}
	\end{figure}
	
	The concept of perceptual loss was first introduced by Johnson et al.~\cite{perceptual} who tried to solve the image transformation problem by comparing content and style discrepancies between two images. They used VGG-16~\cite{vgg} pre-trained for image classification as the loss network and measured perceptual differences of output and ground-truth images. Motivated by their method, we propose a loss function that compares feature differences in the high-frequency domain instead of comparing per-pixel differences in a color space.

	The commonly used perceptual loss uses the VGG-16 network pre-trained on ImageNet dataset. However, the network trained on image classification is optimized to extract feature representation that contains information about the class of objects in images. The objective of network is to figure out what objects are in the image, not to extract detailed patterns or complex high-frequency information. What we need, however, is neural networks that extract such detailed patterns and high-frequency information and feature representations of the networks that are needed to extract such information. Because the commonly used perceptual loss is not appropriate to our problem solving, we have trained a neural network that is optimized to high-frequency extraction.
	
	We first extracted high-frequency signals by applying a high-pass filter to the image transformed into the frequency domain via Fast Fourier Transform. Then, we trained a simple three-layered CNN, or high-pass filtering network, which takes images as input and generates high-pass filtered signals. Figure \ref{fig:highpass_filter} shows an example of high-frequency signals extracted by a traditional high-pass filter using FFT and our high-pass filtering network. Figure \ref{fig:highpass_model} shows a visualization of feature maps produced by intermediate layers of the network during extracting high-frequency signals.

	We utilized this high-pass filtering network as the loss network and defined the high-pass filtering loss function as following:
	\begin{equation}\label{eqn:hfloss}
		\mathcal{L}_{hf}(I_{SR}, I_{HR}) = \mathcal{L}_{0}^{\phi}(I_{SR}, I_{HR}) + \mathcal{L}_{1}^{\phi}(I_{SR}, I_{HR})
	\end{equation}
	where $\phi$ denotes the high-pass filtering network. The loss network $\phi$ analyzes images from various perspectives to generate high-frequency signals where intermediate layers give us abstract feature maps, including edges. We measure the feature difference of $I_{SR}$ and $I_{HR}$ by feed-forwarding two images to fixed $\phi$ where the feature difference is trainable by back-propagation as it is generated through convolutional layers. By minimizing the high-pass filtering loss, high-frequency features are added to the $I_{SR}$, allowing us to obtain sharper images.

	\subsection{Soft Gradient Magnitude Similarity Map Masking}
	
	\begin{figure}[h!]			
		\centering			
		\subfigure[Hard Gradient Magnitude Similarity Map Masking]{			
			\includegraphics[width=.8\columnwidth]{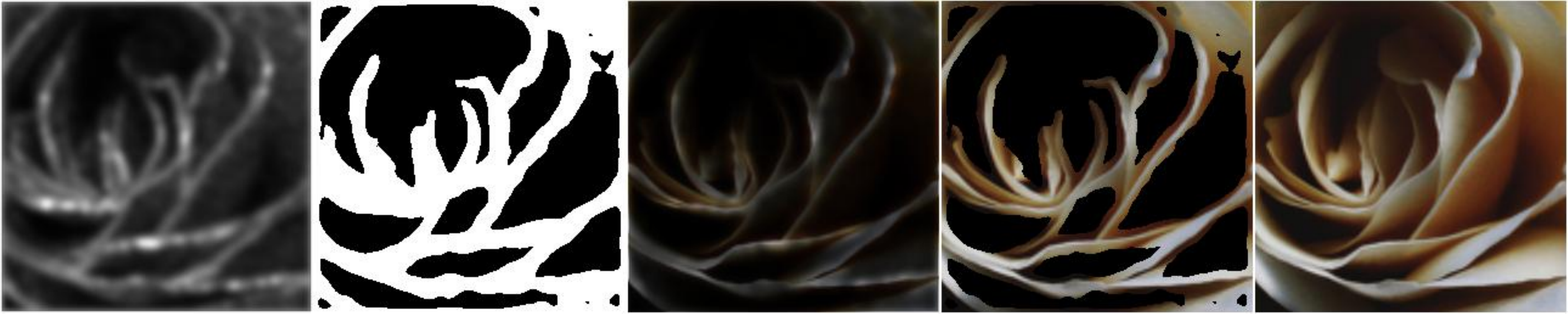}		
			\label{fig:gsmm_hard}				
		}			
		\subfigure[Soft Gradient Magnitude Similarity Map Masking]{				
			\includegraphics[width=.8\columnwidth]{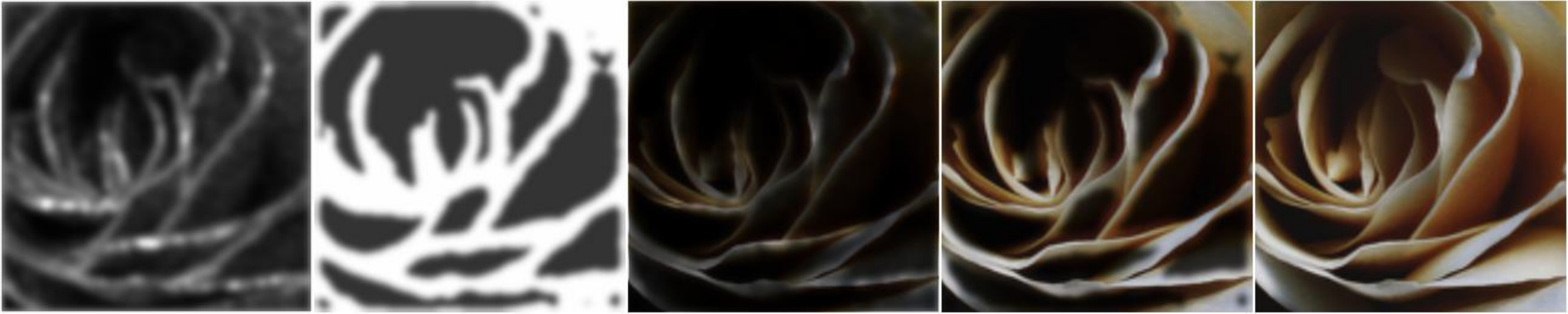}		
			\label{fig:gsmm_soft}
		}
		\caption{Visual examples of hard and soft version of Gradient Magnitude Similarity map masking. From left to right: GMS map, binarized GMS map, GMS map masked image, Hard/Soft GMS map masked image, and Original Image.}
		\label{fig:sgmm}
	\end{figure}

	The local perceptual quality of output images often varies by region. In general, the lower perceptual quality is more observed in areas containing detailed and irregular patterns, but these results depend on which model is used. To learn models that perform evenly, we need to know which part of the resulting image is poor, and therefore more training is needed.
	
	Here, we adopted the Gradient Magnitude Similarity (GMS) map~\cite{gmsd} to evaluate the local quality of images. The gradient magnitude of given image $I$ is computed as follows:
	\begin{align}
		GM(I) = \sqrt{(I\ast G_x)^2 + (I\ast G_y)^2} \label{eqn:gradient_magnitude}
	\end{align}
	where $G_x$ and $G_y$ denote prewitt filters. With the gradient magnitudes of $I_{HR}$ and $I_{SR}$, we compute the GMS map as follows:
	\begin{align}
		GMS(I_{HR}, I_{SR}) = 1 - \frac{2 GM(I_{HR}) GM(I_{SR}) + c}{GM(I_{HR})^2 + GM(I_{SR})^2 + c} \label{eqn:gradient_magnitude_similarity}
	\end{align}
	where we set $c=170$ for pixel values in $[0,255]$. Note that the value of the GMS map is closer to zero where two images are similar while it is closer to one where two images are different.
	
	To give information about which area is more damaged and thus training should be weighted to the loss function, we multiply $I_{HR}$ and $I_{SR}$ with the GMS map before we put them into our loss functions. However, since the GMS map is calculated pixel-wise, it can be computed high for some lucky locations with similar pixel values even where they are contained in severely corrupted regions. So we first binarized the GMS map and then remove tiny or trivial regions using the opening which is defined as erosion followed by dilation. 
	
	In the mathematical morphology, the opening of a binary image $A$ by the structuring element $B$ is expressed as follows:
	
	\begin{align}
		\mbox{\textit{Erosion:}}\quad A \ominus B &= \bigcap_{b\in B}A_{-b} \label{eqn:erosion} \\			
		\mbox{\textit{Dilation:}}\quad A \oplus B &= \bigcup_{b\in B}A_{b} \label{eqn:dilation} \\
		\mbox{\textit{Opening:}}\quad A \circ B &= (A\ominus B)\oplus B \label{eqn:opening}
	\end{align}
	where $A_b$ denotes the translation of $A$ by $b$. The opening is often applied to coarse images to remove pixel-wise outliers and make them locally smooth. Here, adopting the opening to the coarse GMS map allows us to eliminate pixel noise and acquire more smooth labels while maintaining information about the locally damaged area inside the image.
	
	Two images on left side of Figure \ref{fig:gsmm_hard} shows an visual example of applying the opening to GMS map. We can obseved that the map distinguishes between well-reconstructed and poorly-reconstructed area smoother when the opening followed by thresholding is applied to the coarse GMS map. Here, we use the opened-binarized GMS map, or the hard GMS map, to mask images to let our network re-train only on poorly-reconstructed areas.
	
	The hard GMS map assigns each pixel a hard label whether to train or not. In practice, however, it is more reasonable to express with score or probability of how much pixel should be trained. Therefore, we transform the discretized hard GMS map into soft GMS map so it represent the pixel-wise score.
	
	To soften the hard GMS map, we smoothed the boundaries between different regions within the hard GMS map by applying blurring with isotropic Gaussian kernel and additional image opening to remove outliers in an iterative manner. In the soft GMS map, pixels at the center of well or poorly-reconstructed area have more confident score close to 0 or 1, respectively, while scores close to 0.5 are assigned to pixels if they are close to boundaries. Figure~\ref{fig:sgmm} show examples of masked results using the hard and the soft GMS maps.
	
	\subsection{Network Architecture}
	
	Figure~\ref{fig:network} shows the structure of our proposed network. Our deblurring model first resizes image into three different scales, then extracts low-level and high-level features through its head and body for each scale, respectively. The heads of the network consist of one convolutional layer each, like our denoising model, while the bodies are composed of a different number of Residual Channel Attention Blocks depending of the scale; from largest to smallest scale, each body consists of 4, 16, and 64 blocks, respectively. Here, we let bodies on smaller scales have more blocks with deeper layers because they use less GPU memory as their computation is relatively lower.
	
	To take full advantage of deeply stacked bodies on smaller scale, we combined low-level features from smaller scales with those from larger scales through our feature attention module before we put them into each body. This allows deeper bodies for smaller scales take more diverse features and generate richer high-level features.
	
	After the bodies extract high-level features, we combine high-level features with low-level and multi-scale edge filtering module for each scale by feature attention. Here, we upscale and combine high-level features from smaller scale to larger scales, which allows the tails for larger scale take more features from diverse scales. To send features to different scales, we use strided convolution for downscale and pixel shuffle to upscale the features.
	
	\section{Experiment}
	
	%
	%
	
	\subsection{Training Detail}
	
	In the training phase, we trained our model for $800$ epochs for small image patches and $20$ epochs for large image patches with Adam optimizer and initial learning rate $10^{-4}$ with learning rate decay by $0.99$ by every $1,000$ steps. For each iteration, $16$ batches with $192\times 192$ sized cropped patches from large images were used for epochs with small images where one batch with $320\times 180$ sized input image and $1280\times 720$ sized target image was used for epochs with large images. Lastly, L1 function was used for the loss function.
	
	\subsection{Experimental Result}
	
	In this section, we introduce the results of our model applied to Image Denoising, Image Deblurring, and Single Image Super-Resolution.
	
	\subsubsection{Image Denoising}

	We first applied our model to a synthetic noisy dataset generated by adding white Gaussian noise with $\sigma=10$ and $30$ to DIV2K dataset, respectively. In the training phase, we optimized our model for every variance of the noise at once. Using different kinds of noise together, our model has become flexible to more diverse noise levels.
	
	Table~\ref{table:denoise_div2k} shows a comparison of the results of our model and other learning-based models using PSNR and SSIM scores. Our proposed model proved the best performance in most cases.

	Furthermore, we used the Smartphone Image Denoising Dataset, which is often called SIDD~\cite{sidd} to evaluate our model on real noisy images. This dataset consists of pairs of real noisy images taken under various conditions using smartphone cameras and ground-truth images, of which defective pixels are corrected manually. We showed that our model could solve real-world denoising problems by training and evaluating our model on the SIDD dataset.
	
	Table~\ref{table:denoise_sidd} shows a comparison of the results of our model and other learning-based models on the SIDD dataset.

	\begin{figure}[h]
		\centering
		\subfigure[DIV2K]{\includegraphics[width=.45\textwidth]{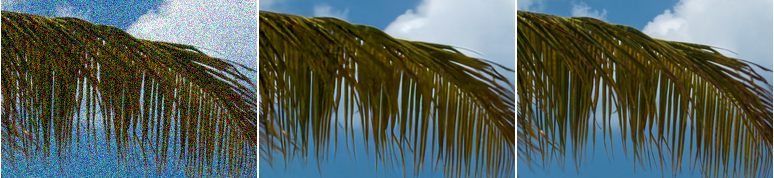}}
		\subfigure[DIV2K]{\includegraphics[width=.45\textwidth]{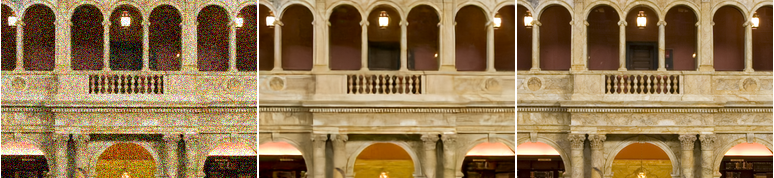}}
		\subfigure[SIDD]{\includegraphics[width=.45\textwidth]{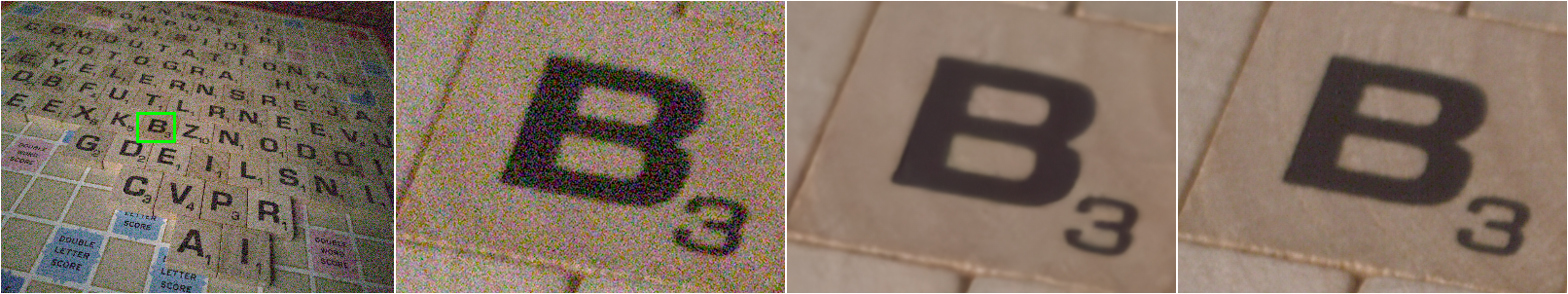}}
		\subfigure[SIDD]{\includegraphics[width=.45\textwidth]{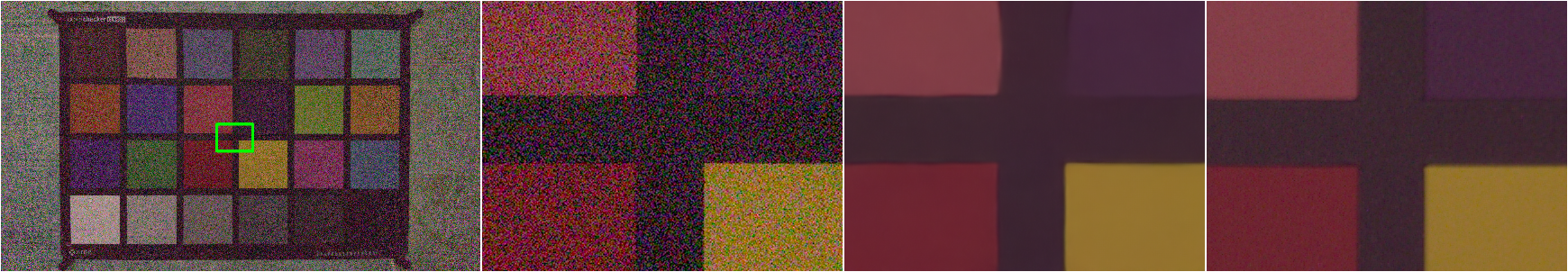}}
		\caption{Our Image Denoising results.}\label{fig:result_denoise}
	\end{figure}
		
	\setlength{\tabcolsep}{7pt}
	\begin{table}[h]
		\newcolumntype{?}{!{\vrule width 1.0pt}}
		\renewcommand{\arraystretch}{1.2}
		\begin{center}
			\caption{Comparison of denoising results in PSNR and SSIM scores on DIV2K + AWGN dataset. Best scores marked in bold.}
			\label{table:denoise_div2k}	
			\begin{tabular}{l|c|c}
				\thickhline
				\multirow{2}{*}{\textbf{Method}} & \multicolumn{2}{c}{\textbf{DIV2K + AWGN}}\\
				\cline{2-3}
				& \multicolumn{1}{c|}{$\sigma=10$} & \multicolumn{1}{c}{$\sigma=30$} \\
				\thickhline
				Noisy Images			&	32.95 / 0.7037	&	23.41 / 0.3280	\\
				DnCNN~\cite{dncnn}		&	30.28 / 0.8753	&	26.74 / 0.6389	\\
				MemNet~\cite{memnet}	&	33.36 / 0.8815	&	29.67 / 0.6619	\\
				FFDNet~\cite{ffdnet}	&	30.06 / 0.8208	&	29.04 / 0.7795	\\
				DHDN~\cite{dhdn}		&	35.31 / 0.8900	&	29.74 / 0.7401	\\
				Ours w/o Edge			&	\textbf{38.64} / 0.9476	&	31.63 / 0.8325	\\
				Ours w/o FeaAtt			&	38.62 / 0.9475	&	31.65 / 0.8313		\\
				Ours					&	\textbf{38.64} / \textbf{0.9483}	
				&	\textbf{31.67} / \textbf{0.8361}	\\
				\thickhline
			\end{tabular}			
		\end{center}
	\end{table}
	\setlength{\tabcolsep}{1.4pt}
	
	\setlength{\tabcolsep}{18pt}
	\begin{table}[h]
		\newcolumntype{?}{!{\vrule width 1.0pt}}
		\renewcommand{\arraystretch}{1.2}
		\begin{center}
			\caption{Comparison of denoising results in PSNR and SSIM scores on SIDD dataset. Best scores marked in bold.}
			\label{table:denoise_sidd}	
			\vspace{1em}
			\begin{tabular}{l|c|c}
				\noalign{\smallskip}
				\thickhline
				\multirow{2}{*}{\textbf{Method}} & \multicolumn{2}{c}{\textbf{SIDD}}\\
				\cline{2-3}
				& \multicolumn{1}{c|}{\textbf{PSNR}} & \multicolumn{1}{c}{\textbf{SSIM}} \\
				\thickhline
				Noisy Images			&	34.19	&	0.5472	\\
				DnCNN~\cite{dncnn}		&	43.60	&	0.9275	\\
				MemNet~\cite{memnet}	&	44.24	&	0.9249	\\
				DHDN~\cite{dhdn}		&	46.99	&	0.9677	\\
				Ours w/o Edge			&	47.14	&	0.9692	\\
				Ours w/o FeaAtt			&	47.12	&	\textbf{0.9693}  \\
				Ours					&	\textbf{47.21}	&	\textbf{0.9693}  \\
				\thickhline
			\end{tabular}			
		\end{center}
	\end{table}
	\setlength{\tabcolsep}{1.4pt}
	
	\begin{figure}[h]
		\centering
		\subfigure[DIV2K]{\includegraphics[width=.45\textwidth]{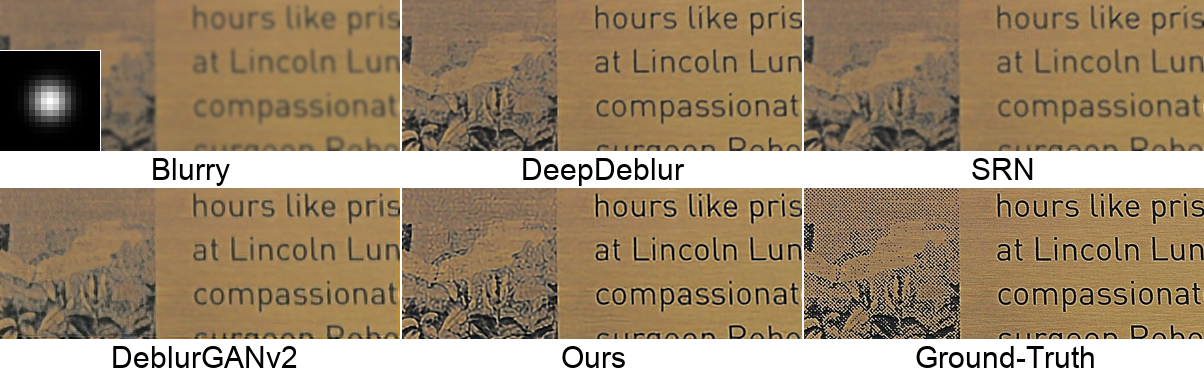}}
		\subfigure[REDS]{\includegraphics[width=.45\textwidth]{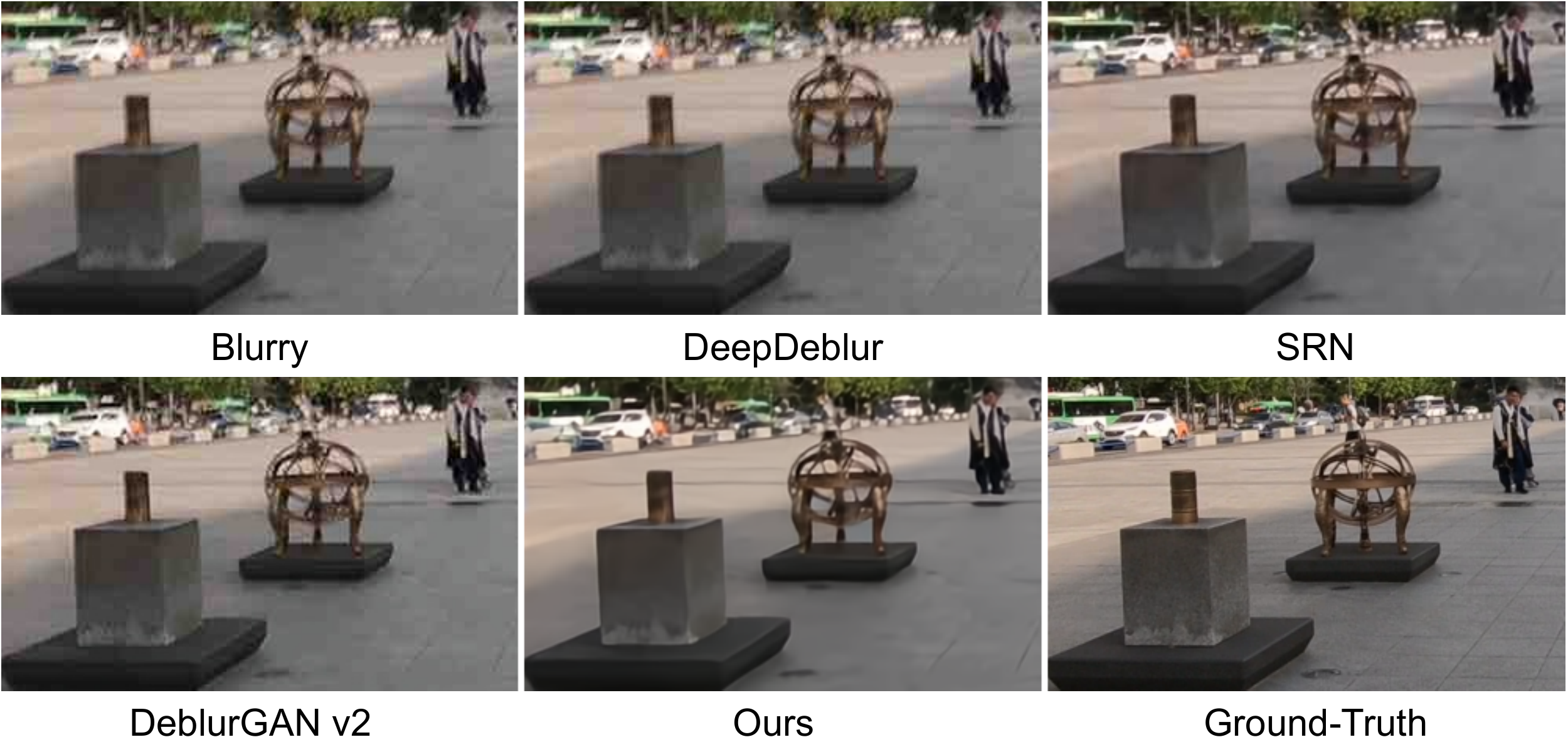}}
		\caption{Our Image Deblurring results.}\label{fig:result_deblur}
	\end{figure}
	
	\begin{figure}[h]
		\centering
		\subfigure[LR image]{\includegraphics[width=.1\textwidth]{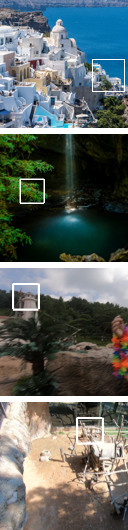}}
		\subfigure[L1 trained]{\includegraphics[width=.1\textwidth]{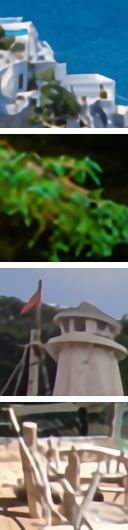}}
		\subfigure[Perceptual]{\includegraphics[width=.1\textwidth]{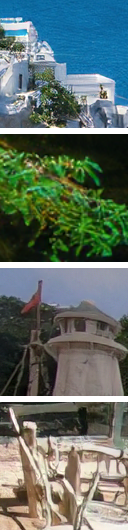}}
		\subfigure[HR image]{\includegraphics[width=.1\textwidth]{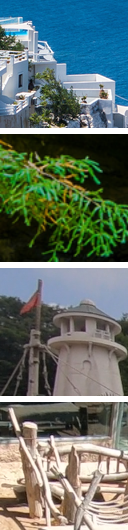}}
		\caption{Visual comparison of PSNR-oriented and Perceptual methods}\label{fig:result_f}
	\end{figure}

	\subsubsection{Image Deblurring}
	
	We first trained and evaluated our deblurring model on Flickr2K dataset~\cite{edsr}. While REDS consists of similar images from several daily videos, Flickr2K contains different images with various objects and detailed patterns. Therefore, it is suitable for an extensive experiment to show that our model can be applied to images with more diverse information.
	
	To create blurry images in various conditions, we applied randomly chosen blur kernels from set of isotropic and anisotropic Gaussian kernels of various sizes and angles to randomly cropped and rotated patches. By augmenting the blurry images, our model could observe and learn the various blurring conditions on the limited images.
	
	Table~\ref{table:deblur_flickr} shows our model achieves the state-of-the-art results on Flickr2K dataset.

	Table~\ref{table:deblur} shows our experimental results on the REDS dataset from ``NTIRE 2021 Image Deblurring Challenge - Track2. JPEG Artifacts''.
	
	We also provide the results of ablation studies that evaluate the effect of our Multi-Scale Edge Filtering and Feature Attention Module. Ablations studies show that our proposed model achieves top scores at PSNR, and SSIM. Considering that PSNR measures absolute errors and SSIM measures the perceived change in structural information, based on luminance and contrast of images, it can be inferred that the Feature Attention Module helps the model understand the structural information.
	
	Figure~\ref{fig:result_deblur} shows some selected deblurring results of our proposed model on REDS dataset. Our model successfully reconstruct objects that are difficult to identify from the blurry image to identifiable levels.

	\setlength{\tabcolsep}{17pt}
	\begin{table}[!h]
		\newcolumntype{?}{!{\vrule width 1.0pt}}
		\renewcommand{\arraystretch}{1.2}
		\begin{center}
			\caption{Comparison of deblurring results in PSNR and SSIM scores on Flickr2K dataset. Best scores marked in bold.}
			\label{table:deblur_flickr}	
			\begin{tabular}{l|c|c}
				\thickhline
				\multirow{2}{*}{\textbf{Method}} & \multicolumn{2}{c}{\textbf{Flickr2K}} \\
				\cline{2-3}
				& \multicolumn{1}{c|}{\textbf{PSNR}} & \multicolumn{1}{c}{\textbf{SSIM}} \\
				\thickhline
				Blurry Images			&	29.19	&	0.7655	\\
				DeepDeblur
				~\cite{deepdeblur}		&	33.75	&	0.8990	\\
				SRN~\cite{srn}			&	34.90	&	0.9070	\\
				DeblurGANv2
				~\cite{deblurgan_v2}	&	30.78	&	0.8546	\\
				Ours w/o Edge           &   36.32	&	0.9253	\\
				Ours w/o FeaAtt         &   36.36	&	0.9252	\\
				Ours					&	\textbf{36.38}	&	\textbf{0.9264}	\\
				\thickhline
			\end{tabular}			
		\end{center}
	\end{table}
	\setlength{\tabcolsep}{1.4pt}
	
	\setlength{\tabcolsep}{17pt}
	\begin{table}[!h]
		\newcolumntype{?}{!{\vrule width 1.0pt}}
		\renewcommand{\arraystretch}{1.2}
		\begin{center}
			\caption{Comparison of deblurring results in PSNR and SSIM scores on REDS - JPEG dataset. Best scores marked in bold.}
			\label{table:deblur}
			\begin{tabular}{l|c|c}
				\thickhline
				\multirow{2}{*}{\textbf{Method}} & \multicolumn{2}{c}{\textbf{REDS}} \\
				\cline{2-3}
				& \multicolumn{1}{c|}{\textbf{PSNR}} & \multicolumn{1}{c}{\textbf{SSIM}} \\
				\thickhline				
				Blurry Images			&	26.51	&	0.7063	\\
				DeepDeblur
				~\cite{deepdeblur}		&	28.68	&	0.7690	\\
				SRN~\cite{srn}			&	28.60	&	0.7588	\\
				DeblurGANv2
				~\cite{deblurgan_v2}	&	26.82	&	0.7220	\\
				Ours w/o Edge           &   29.56	&	0.7901	\\
				Ours w/o FeaAtt         &   29.78	&	0.7959	\\
				Ours					&	\textbf{29.80}	&	\textbf{0.7966}	\\
				\thickhline
			\end{tabular}			
		\end{center}
	\end{table}
	\setlength{\tabcolsep}{1.4pt}
	
	\subsubsection{Single Image Super-Resolution}
	
	Compared to other blind SISR methods, our proposed net achieves higher PSNR and SSIMS scores while our proposed GAN produces perceptually more natural results. Figure \ref{fig:result_f} shows detailed results of our models.

	\setlength{\tabcolsep}{8pt}
	\begin{table}[H]
		\renewcommand{\arraystretch}{1.2}
		\begin{center}
			\caption{Comparison of SISR results on DIV2K dataset. Best scores marked in bold.}
			\label{table:sisr_reds}
			\footnotesize
			\begin{tabular}{l|c|c|c}
				\thickhline
				\multirow{2}{*}{\textbf{Method}} & \multicolumn{3}{c}{\textbf{DIV2K}}\\
				\cline{2-4}
				& \multicolumn{1}{c|}{\quad\textbf{PSNR} $\uparrow$}
				& \multicolumn{1}{c|}{\quad\textbf{SSIM} $\uparrow$}
				& \multicolumn{1}{c}{\quad\textbf{LPIPS} $\downarrow$} \\
				\thickhline
				Bicubic				&	26.78	&	0.6839	&	0.4087	\\
				EDSR~\cite{edsr}	&	28.65	&	0.7594	&	0.2451	\\
				RCAN~\cite{rcan}	&	28.93	&	0.7680	&	0.2371	\\
				DRN~\cite{drn}		&	28.91	&	0.7676	&	0.2363	\\
				Our Net	w/o Edge 			&	29.00 & 0.7705 & 0.2240	\\
				Our Net	w/o FeaAtt 			&	29.10 & 0.7740 & 0.2239	\\
				Our Net	&	\textbf{29.11}	&	\textbf{0.7743}	&	0.2225	\\
				Our GAN	&	25.79	&	0.6598	&	\textbf{0.1096}	\\
				
				\thickhline
			\end{tabular}
			\normalsize
		\end{center}
	\end{table}
	
	\setlength{\tabcolsep}{8pt}
	\begin{table}[H]
		\renewcommand{\arraystretch}{1.2}
		\begin{center}
			\caption{Comparison of SISR results on REDS dataset. Best scores marked in bold.}
			\label{table:sisr_reds}	
			\footnotesize
			\begin{tabular}{l|c|c|c}
				\thickhline
				\multirow{2}{*}{\textbf{Method}} & \multicolumn{3}{c}{\textbf{REDS - NTIRE 2021 Low Resolution}}\\
				\cline{2-4}
				& \multicolumn{1}{c|}{\quad\textbf{PSNR} $\uparrow$}
				& \multicolumn{1}{c|}{\quad\textbf{SSIM} $\uparrow$}
				& \multicolumn{1}{c}{\quad\textbf{LPIPS} $\downarrow$} \\
				\thickhline
				Bicubic	&	24.55	&	0.6313	&	0.4855	\\
				EDSR~\cite{edsr}	&	25.26	&	0.6775	&	0.3752	\\
				RCAN~\cite{rcan}	&	25.31	&	0.6797	&	0.3775	\\
				DRN~\cite{drn}	&	25.30	&	0.6791  &	0.3773	\\
				Our Net w/o Edge	&	27.78	&	0.7648  &	0.2542	\\
				Our Net w/o FeaAtt	&	\textbf{27.83}	&	0.7660  &	0.2550	\\
				Our Net	&	\textbf{27.83}	&	\textbf{0.7662}	&	0.2540	\\
				Our GAN	&	24.19	&	0.6337	&	\textbf{0.1489}	\\
				\thickhline
			\end{tabular}	
			\normalsize		
		\end{center}
	\end{table}

	\section{Conclusion}
	
	This paper introduces multi-scale edge filtering that extracts high-frequency information from noisy images. This helps our model perform statistical analysis and reconstruction suitable for each region of the image. We also add feature attention modules to enable the network to determine feature maps containing more important information. Also, we introduce a high-pass filtering loss function that compares feature maps generated from a high-pass filtering network computing the high-frequency information of the results and ground truth images. Finally, our soft GMS masking helps the model identify which areas of the resulting image are more compromised and need to be more focused on additional training processes.
	
	Experimental results show that our model can achieve state-of-the-art PSNR and SSIM scores compared to other learning-based methods. However, when visualizing the results, over-smoothing problems have been observed as in other PSNR-oriented methods. Adversarial training was applied to pre-trained models using a discriminator that distinguishes real and synthetic images, allowing the model to generate much more natural images.
	
	In future research, we will study learning-based methods that achieve superior scores in both PSNR and LPIPS by enabling the model to extract sufficient information from low-resolution images.




	
	
	%
	
	\bibliographystyle{IEEEtran}
	\bibliography{mybib}
%
%

\end{document}